\pdfoutput=1

\documentclass[11pt]{article}

\usepackage{ACL2023}
\usepackage{hyperref}

\usepackage{times}
\usepackage{latexsym}
\usepackage{booktabs}
\usepackage{float}
\usepackage{graphicx}
\usepackage{amsmath}
\usepackage{fontawesome5}
\usepackage{placeins}

\usepackage[T1]{fontenc}

\usepackage[utf8]{inputenc}

\usepackage{microtype}

\usepackage{inconsolata}

\title{ActiveAED: A Human in the Loop Improves Annotation Error Detection}

\author{Leon Weber\textsuperscript{\faMountain} \and
Barbara Plank\textsuperscript{\faMountain}\textsuperscript{$\diamondsuit$} \\
\textsuperscript{\faMountain}Center for Information and Language Processing (CIS), LMU Munich, Germany \\
\textsuperscript{$\diamondsuit$}Munich Center for Machine Learning (MCML), Munich, Germany\\
{\tt \{leonweber, bplank\}@cis.lmu.de  } }

\begin{document}
\maketitle
\begin{abstract}
  Manually annotated datasets are crucial for training and evaluating Natural Language Processing models. However, recent work has discovered that even widely-used benchmark datasets contain a substantial number of erroneous annotations. This problem has been addressed with Annotation Error Detection (AED) models, which can flag such errors for human re-annotation. However, even though many of these AED methods assume a final curation step in which a human annotator decides whether the annotation is erroneous, they have been developed as static models without any human-in-the-loop component. In this work, we propose ActiveAED, an AED method that can detect errors more accurately by repeatedly querying a human for error corrections in its prediction loop. We evaluate ActiveAED on eight datasets spanning five different tasks and find that it leads to improvements over the state of the art on seven of them, with gains of up to six percentage points in average precision. 
\end{abstract}

\section{Introduction}

Correct labels are crucial for model training and evaluation. 
Wrongly labelled instances in the training data hamper model performance~\citep{larsonInconsistenciesCrowdsourcedSlotFilling2020a,vlachosActiveAnnotation2006}, whereas errors in the test data can lead to wrong estimates of model performance~\citep{altTACREDRevisitedThorough2020,larsonInconsistenciesCrowdsourcedSlotFilling2020a,reissIdentifyingIncorrectLabels2020}.
This is a problem in practice, as even widely used benchmark datasets can contain a non-negligible number of erroneous annotations~\citep{altTACREDRevisitedThorough2020,northcuttPervasiveLabelErrors2021,reissIdentifyingIncorrectLabels2020}.
Researchers have developed a multitude of annotation error detection (AED) methods to detect such labelling errors as recently surveyed by \citet{klieAnnotationErrorDetection2022}.
After detection, there are multiple ways to deal with the found annotation errors.
When it comes to training data, a reasonable strategy is to simply remove the instances flagged by an AED model~\citep{huangO2UNetSimpleNoisy2019}.
For evaluation data, however, this is not viable, because in many cases this would remove a significant fraction of hard but correctly labelled instances in addition to the errors~\citep{swayamdiptaDatasetCartographyMapping2020}, which would lead to an overestimation of model performance.
Instead, researchers resorted to manual correction of the labels flagged by the AED method~\citep{altTACREDRevisitedThorough2020,reissIdentifyingIncorrectLabels2020,northcuttPervasiveLabelErrors2021,larsonInconsistenciesCrowdsourcedSlotFilling2020a}.
Strikingly, even though this manual correction requires human input, the typical workflow is to first apply the AED method once and afterwards correct the flagged errors, without using the human feedback in the AED step.

\begin{figure}[t]
  \centering
  \includegraphics[width=\linewidth]{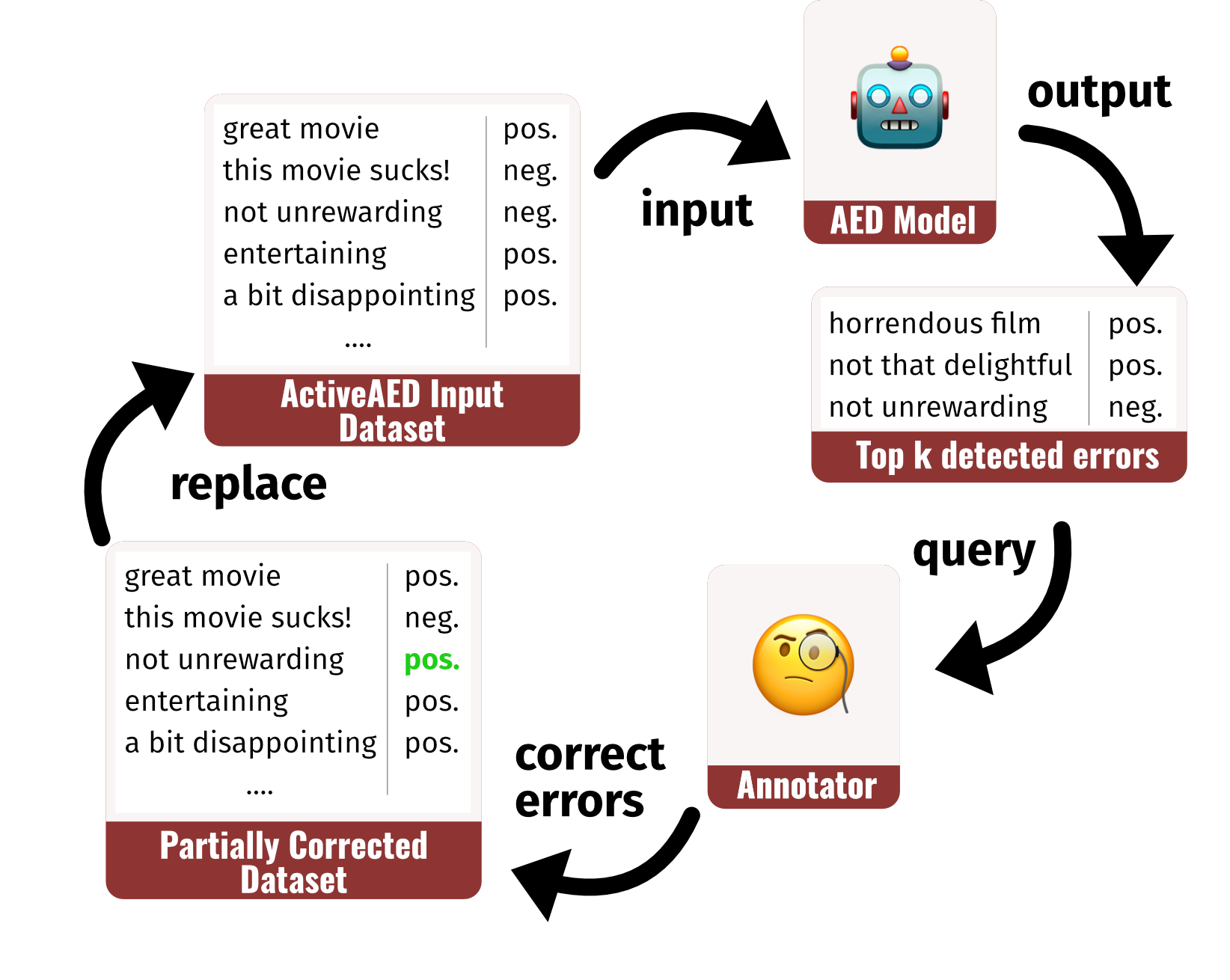}
  \caption{Prediction loop of ActiveAED}
  \label{fig:activeaed}
\end{figure}

We hypothesize that connecting the human input and the AED prediction in a human-in-the-loop setup could increase the accuracy of the AED method without increasing the total amount of human intervention.
To support this hypothesis, we propose ActiveAED, an AED method which includes human feedback in the annotation loop; see Figure~\ref{fig:activeaed} for an illustration.
We base ActiveAED on the Area-under-the-Margin metric (AUM)~\citep{pleissIdentifyingMislabeledData2020}, which was recently proposed to detect annotation errors in computer vision datasets.
As an additional contribution, we propose a novel ensembling scheme to improve AUM's performance.
In experiments on eight datasets spanning five different tasks, we show that ActiveAED improves over three baselines that performed well in a recent evaluation~\citep{klieAnnotationErrorDetection2022}.
On seven datasets, we observe improvements, with gains of up to six percentage points (pp) in average precision. 
Our ablation study shows that both the human-in-the-loop component and the ensembling scheme contribute to the improvements.
We make code and data available under \url{https://github.com/mainlp/ActiveAED}.

\section{Related Work}
AED for Natural Language Processing (NLP) datasets has a long tradition which has recently been comprehensively evaluated and surveyed by the seminal work of \citet{klieAnnotationErrorDetection2022}.
We base our evaluation setup on theirs.
Existing AED methods can be divided into six different categories~\citep{klieAnnotationErrorDetection2022}: variation-based~\citep{dickinsonDetectingErrorsPartofSpeech2003,larsonInconsistenciesCrowdsourcedSlotFilling2020a},
model-based~\citep{amiriSpottingSpuriousData2018,yaghoub-zadeh-fardStudyIncorrectParaphrases2019,chong-etal-2022-detecting},
training-dynamics-based~\citep{swayamdiptaDatasetCartographyMapping2020,pleissIdentifyingMislabeledData2020,siddiquiMetadataArchaeologyUnearthing2022}, vector-space-proximity-based~\citep{larsonOutlierDetectionImproved2019,grivasNotCuteStroke2020}, ensembling-based~\citep{altTACREDRevisitedThorough2020,varshneyILDAEInstanceLevelDifficulty2022} and rule-based~\citep{kvetonSemiAutomaticDetection2002}.
To the best of our knowledge, none of these AED methods has been developed or evaluated with a human in the loop, except for \citet{vlachosActiveAnnotation2006} who uses AED as part a larger framework for constructing a silver-standard dataset.
Accordingly, they do not compare the performance of the AED component to competing approaches and they consider only a single dataset and task.

Additionally, one can distinguish between flaggers and scorers for AED~\citep{klieAnnotationErrorDetection2022}.
Flaggers output hard decisions of whether an instance contains an error, whereas scorers assign to each instance a score reflecting the likelihood of being an error. 
In this work, we focus on scoring methods, because ActiveAED requires error scores to rank the instances.

\section{Active Annotation Error Detection}
We propose ActiveAED, an AED method which uses the error corrections issued by an annotator in its prediction loop.
The basic procedure of ActiveAED is this:
In the first step, it uses a ranking-based AED method to find the $k$ most likely annotation errors across the dataset.
In the second step, the presumed annotation errors are forwarded to an annotator who checks them and corrects the labels if necessary.
After this, the dataset is updated with the corrections issued by the annotator and the procedure continues with the first step.
This loop continues until a stopping condition is met, e.g. that the fraction of errors in the batch drops to a user-defined threshold.
See Figure~\ref{fig:activeaed} for an illustration of the process.
We consider a scenario where an annotator wants to correct annotation errors in a dataset with a given annotation budget of $n$ instances.
There are two options of how to apply an annotation error detection (AED) method to support this.
The first is the state-of-the-art and the second one is our proposed approach:
(1) Run the AED method once on the dataset to retrieve a list of instances ranked by their probability of containing an annotation error.
Then, spend the annotation budget by correcting the top-$n$ instances.
(2) Run the AED method and spend some of the annotation budget by correcting the top-$k$ instances with $k$ << $n$.
Then, run the AED method again on the now partially corrected dataset and repeat until the annotation budget is exhausted.
Note, both approaches involve ranking instances based on their probability of containing annotation errors, and selection of a subset of instances for annotation based on this ranking.
As a result, the outputs of both approaches can be fairly compared, because they use the same annotation budget and the same ranking-based score.

More formally, we assume a dataset with inputs $X$, (potentially erroneous) labels $y$, and true labels $y^*$ which are initially unknown to us.
After training the model for $E$ epochs, we use (negative) AUM to assign error scores:
\begin{align}
  s_{i} = \frac{1}{E} \sum_{e=1}^{E} \max_{y' \neq y_i} p_{\theta_{e}}(y'| x_i) - p_{\theta_{e}}(y_i | x_i),
  \label{eq:aum}
\end{align}
where $p_{\theta_{e}}(y_i | x_i)$ is the probability of the label assigned to $x_i$ as estimated by $\theta_{e}$ and $\max_{y' \neq y_i} p_{\theta_{e}}(y'| x_i)$ the probability of the highest scoring label that is not the assigned one.
Intuitively, correctly labelled instances on average obtain smaller (negative) AUM scores (Eq.~\ref{eq:aum}) than incorrect ones, because the model will confidently predict their correct label earlier in the training.
We chose AUM, because it performed well in preliminary experiments on SI-Flights~\citep{larsonInconsistenciesCrowdsourcedSlotFilling2020a} and ATIS~\citep{hemphillATISSpokenLanguage1990}.
Note, that this formulation differs from the original one in \citet{pleissIdentifyingMislabeledData2020} that uses raw logits instead of probabilities.
We chose to use probabilities because this performed better in our experiments (see Table~\ref{tab:results}).

We extend AUM with a novel ensembling scheme based on training dynamics.
For this, we train a model for $E$ epochs in a $C$-fold cross-validation setup.
For each fold $c \in \{1, ..., C\}$ and  epoch $e \in \{1, ..., E\}$, we obtain a model $\theta_{c,e}$. 
We use the models of one fold $c$ to assign an error score $s_{c,i}$ to each instance with AUM (Eq.~\ref{eq:aum}).
For each fold, we calculate the AUM score both on the train and on the test portion of the fold, which yields $C-1$ training-based scores and one test-based score for each instance.
For each instance, we first average the training-based scores and then compute the mean of this average and the test-based score, which results in the final score $s_i$:
\begin{align}
  s^{\mbox{\scriptsize \textit{train}}}_i &= \frac{1}{E-1} \sum_{c \in \textit{train}_i} s_{c,i} \\
  s_i &= \frac{1}{2} ( s^{\mbox{\scriptsize \textit{train}}}_i + s^{\mbox{\scriptsize \textit{test}}}_i  ),
\end{align}
where $\textit{train}_i$ is the set of $C-1$ folds in which instance $i$ appears in the training portion. 
Then, we rank all uncorrected instances by $s_i$ and route the $k$ highest scoring ones to the annotator, who manually corrects their label by setting $y_i := y^*_i$.
Finally, the procedure continues with the partially corrected dataset until a stopping condition is met.
There are two kinds of motivation for the proposed ensembling scheme: $s^{\mbox{\scriptsize \textit{train}}}$ should improve the calibration of the model~\citep{ovadiaCanYouTrust2019}, which \citet{klieAnnotationErrorDetection2022} show to be helpful for AED.
$s^{\mbox{\scriptsize \textit{test}}}$ derives from the observation that model-based AED methods benefit from computing statistics over unseen data~\citep{klieAnnotationErrorDetection2022}.

\section{Evaluation Protocol}
\begin{table*}
  \centering
  \small
    \begin{tabular}{l|ll|llllll}
          \toprule
          & ATIS  & SI-Flights & IMDb  & SST   & GUM   & CONLL-2003 & SI-Companies & SI-Forex \\
          \midrule
    CU    & 91.7±1.4 & 80.9±0.5 & 31.6±1.3 & 42.7±1.0 & 98.8±0.1 & 25.2±0.6 & 96.1±0.2 & 84.2±2.0 \\
    DM    & 97.2±0.2 & 79.2±2.4 & 30.1±3.0 & 47.1±1.0 & 99.3±0.1 & 30.2±0.7 & 97.5±0.2 & 80.6±0.9 \\
    AUM (p)   & 98.0±0.1 & 78.9±2.3 & 30.1±3.0 & 47.1±1.0 & 99.0±0.1 & 30.2±0.7 & 97.3±0.3 & 81.1±0.9 \\
    AUM (l)  & 97.3±0.4 & 72.6±0.3 & 27.5±2.5  & 39.6±1.3 & \textbf{99.5±0.1} & 29.3±0.2 & 97.2±0.2 & 66.6±1.5 \\
    ActiveAED & \textbf{98.6±0.1} & \textbf{86.6±0.5} & \textbf{36.6±0.1} & \textbf{53.0±0.2} & 98.5±0.0 & \textbf{33.3±0.2} & \textbf{99.3±0.0} & \textbf{89.7±0.6} \\
    \midrule
    w/o active & 98.7±0.1 & 80.3±0.6 & 36.0±0.4 & 52.9±0.4 & 98.4±0.0 & 31.7±0.4 & 97.9±0.1 & 85.5±0.6 \\
    \bottomrule
    \end{tabular}%
  \caption{Evaluation results. All scores are mean and standard deviation of AP for AED in percent over three random seeds. The best score per dataset (without ablation) is in bold. We used ATIS and SI-Flights as development data. The last row is ActiveAED without the human-in-the-loop component. AUM (l) is the original version of AUM proposed by \citet{pleissIdentifyingMislabeledData2020}, whereas AUM (p) is our variant in which we aggregate probabilities instead of raw logits.}	
  \label{tab:results}%
\end{table*}%

\subsection{Datasets \& Evaluation Setting}
We evaluate ActiveAED on eight datasets following the choice of datasets used by \citet{klieAnnotationErrorDetection2022}:\footnote{From this list, we exclude \citet{plankLinguisticallyDebatableJust2014} because it contains only annotation ambiguities and not corrected errors which are required for our evaluation setting.}
\begin{itemize}
  \item The intent classification part of \textbf{ATIS}~\citep{hemphillATISSpokenLanguage1990}, for which we randomly perturb labels.
  \item The sentiment analysis dataset \textbf{IMDb}~\citep{maas-EtAl:2011:ACL-HLT2011}, for which \citet{northcuttPervasiveLabelErrors2021} provide semi-automatically detected annotation errors.
  \item The sentiment analysis dataset \textbf{SST}~\citep{socher-etal-2013-recursive} with randomly perturbed labels.
  \item The UPOS annotations\footnote{\url{https://github.com/UniversalDependencies/UD_English-GUM}} from the Georgetown University Multilayer Corpus (\textbf{GUM}; \citet{Zeldes2017}) with randomly perturbed labels.
  \item The \textbf{CoNLL-2003} Named Entity Recognition data~\citep{tjongkimsangIntroductionCoNLL2003Shared2003}, for which \citet{reissIdentifyingIncorrectLabels2020} provide a version with corrected annotations.
  \item The slot three filling datasets \textbf{SI Companies}, \textbf{SI Flights}, and \textbf{SI Forex}~\citep{larsonInconsistenciesCrowdsourcedSlotFilling2020a} that contain manually corrected slot labels.
\end{itemize}
We provide Hugging Face datasets implementations and detailed statistics for all datasets; see Appendix~\ref{app:datasets}.
Our evaluation setup for the sequence labelling datasets (GUM, CoNLL-2003, SI Companies, SI Flights, and SI Forex) differs from that proposed by \citet{klieAnnotationErrorDetection2022}. 
We opt for a sequence-level setting because it is closer to our envisioned application scenario, as it makes more sense for an annotator to correct the entire sequence of annotations instead of a single one at a time.
Specifically, we define errors on the sequence level, i.e.\ if at least one token annotation differs from the gold annotation, the sequence is treated as an error both during ActiveAED prediction and for evaluation.
During prediction, ActiveAED aggregates token-level error scores by calculating the maximum over all tokens in the sequence.
For the other parts of the evaluation setup we follow \citet{klieAnnotationErrorDetection2022}.\footnote{Note, that our results are not comparable with the numbers for the state-of-the-art reported by \citet{klieAnnotationErrorDetection2022}, because of the different treatment of sequence-labelling datasets. Additionally, for ATIS and SST the choice of randomly perturbed labels differs (but the fraction is the same) and for IMDb the dataset statistics reported by \citet{klieAnnotationErrorDetection2022} are different from those of the original dataset~\citep{northcuttPervasiveLabelErrors2021}, which we use.}

In all datasets in which we perturbed labels, we resample the label uniformly for 5\% of all annotations.
We use average precision (AP) as our evaluation metric, which we compute with scikit-learn v1.1.3~\citep{scikit-learn}.
To be consistent with ActiveAED's application scenario, we cannot use the standard train/dev/test split practice from supervised learning, because we will not have access to any known errors which we could use for development when we apply ActiveAED to a new dataset.
Thus, we select the two datasets ATIS and SI-Flights as development datasets on which we devise our method, and reserve the remaining datasets for the final evaluation.
We report the average and standard deviation across three random seeds.
We follow the standard practice in active learning research and simulate the annotator by using gold-standard corrections~\citep{settlesActiveLearning2012,zhangALLSHActiveLearning2022}.
Note, that here, we simulate a single annotator without accounting for inter- and intra-annotator variation~\citep{jiangInvestigatingReasonsDisagreement2022,plank-2022-problem}.
We set $k = 50$ (an ablation for $k$ can be found in Section~\ref{sec:results}), because this is small enough so that an annotator can handle it in a single annotation session but large enough that gains can be observed after a single iteration on SI Flights.
We stop the prediction loop after 40 iterations or when the whole dataset was annotated.
We perform 10-fold cross validation in all experiments.
We describe the remaining hyperparameters in Appendix~\ref{app:hyperparameters}.

\subsection{Baselines}
As baselines, we choose the top-performing scorer methods recommended by \citet{klieAnnotationErrorDetection2022}:
\begin{itemize}
  \item (Negative) Area-under-the-margin (\textbf{AUM}) \citep{pleissIdentifyingMislabeledData2020}: $s^{\mbox{\scriptsize\textit{AUM}}}_{i} = \frac{1}{E} \sum_{e=1}^{E} \max_{y' \neq y_i} p_{\theta_{e}}(y'| x_i) - p_{\theta_{ce}}(y_i | x_i)$
  \item (Negative) Data Map Confidence (\textbf{DM}) \citep{swayamdiptaDatasetCartographyMapping2020}: $s^{\mbox{\scriptsize\textit{DM}}}_i = - \frac{1}{E} \sum_{e=1}^{E} p_{\theta^{(e)}}(y_i | x_i)$
  \item Classification Uncertainty (\textbf{CU}) \citep{klieAnnotationErrorDetection2022}: $s^{\mbox{\scriptsize\textit{CU}}}_i = - p_{\theta^*}(y_i | x_i)$,
\end{itemize}
where AUM and DM are both computed over a single training run and CU is computed with cross-validation over the test portions using the model $\theta^*$ achieving the lowest test loss for the given fold.

\section{Results}
\label{sec:results}

The results of our evaluation can be found in Table~\ref{tab:results}.
ActiveAED outperforms the three baselines on seven of the eight datasets, with gains ranging from 0.6 to 6 pp AP.
We observe a large variance of the AP scores across different datasets, which is in concordance with the findings of \citet{klieAnnotationErrorDetection2022}.
We suspect that the relatively low scores on IMDb and CoNLL-2003 are because the errors were manually annotated after automatic filtering and thus are limited by the recall of the filtering method.
We disentangle the contribution of our proposed ensembling strategy from that of the human-in-the-loop component by ablating the human-in-the-loop (last row in Table~\ref{tab:results}).
We find that on four of the eight datasets, the ensembling alone improves results, whereas on SI Companies, SI Flights, and SI Forex, the main driver for improvements is the human-in-the-loop component.
Generally, the human-in-the-loop component improves over the non-active variant on seven out of eight datasets.
A natural question that arises is whether the human-in-the-loop procedure of ActiveAED can also improve AED methods other than our modified version of AUM.
To investigate this, we evaluate unmodified versions of (negative) AUM and DM on SI Flights and ATIS with our human-in-the-loop setup.
We find that, for SI Flights, AUM/DM improves by 7.4/6.9 pp AP, whereas for ATIS, DM improves by 0.8 pp and AUM's result diminishes by 0.2 pp.
This suggests that a human in the loop might not be helpful for all combinations of datasets and methods, but that it has the potential to significantly improve results for other methods than ActiveAED.

\begin{figure}
  \centering
  \includegraphics*[width=\linewidth]{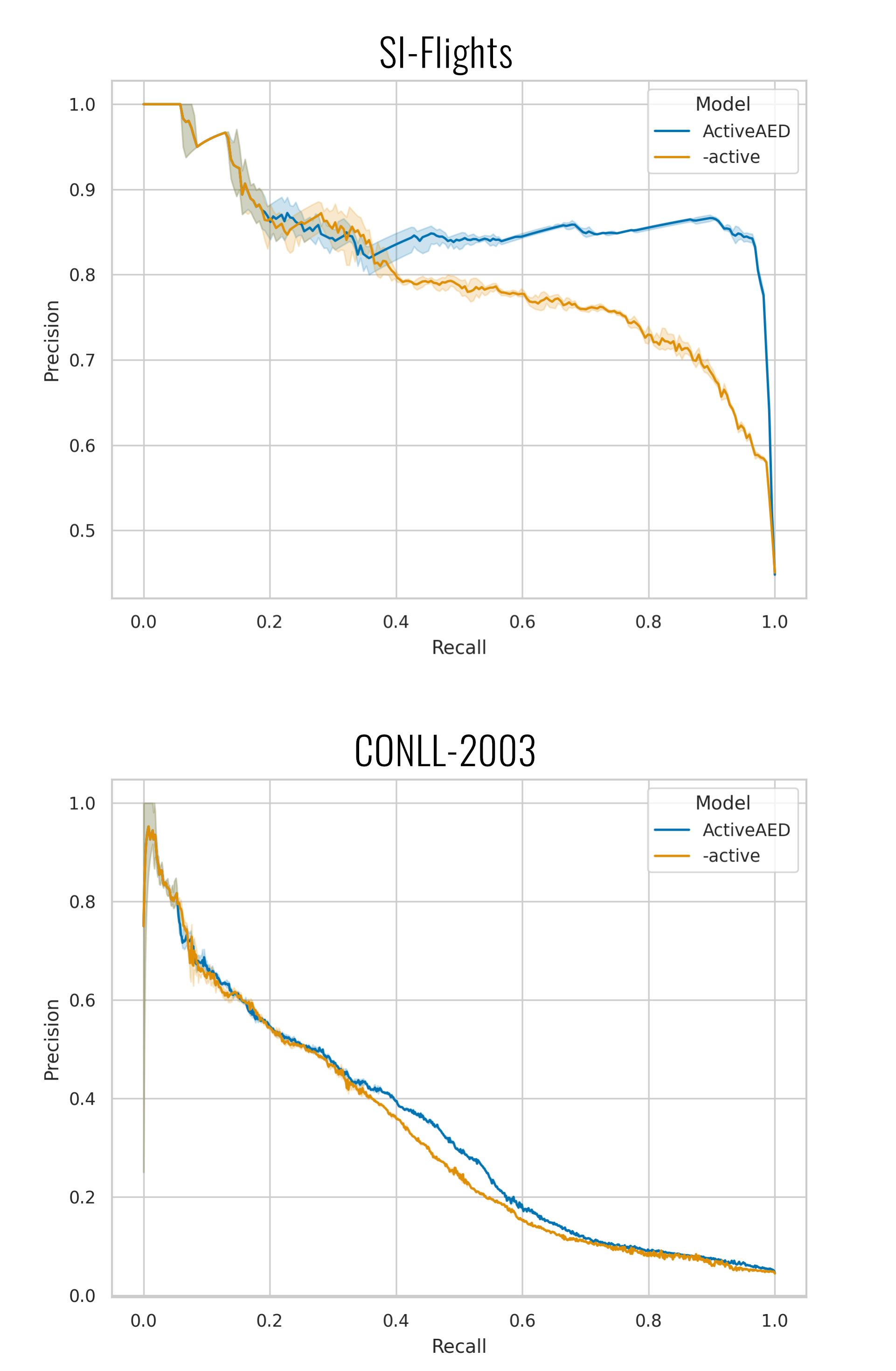}
  \caption{Comparison of the precision-recall curves of ActiveAED and its non-active ablation. The gains of ActiveAED are made in the mid-to-high recall regime for both datasets. Curves are mean and error bars are standard deviation across three random seeds.}
  \label{fig:curves}
\end{figure}

It is instructive to compare the precision-recall curves of ActiveAED to that of its non-active variant.
The graphs for datasets SI Flights and CoNLL-2003 can be found in Figure~\ref{fig:curves}.
On both datasets, the precision gains are present in the mid-to-high recall regime (> 0.4), which intuitively makes sense, because ActiveAED requires a few rounds of human annotation to produce different outputs than its non-active variant.
This suggests that one could increase the efficiency of ActiveAED by starting with a more lightweight AED method, e.g. one that does not require cross validation or ensembling and only later switch to the more compute-intensive ensembling of ActiveAED.
We leave the investigation of this option for future work.
We describe the ablation study of our proposed ensembling scheme and for different choices of $k$ in Appendix~\ref{app:ablation}.
Here, we find that test ensembling is crucial, that train ensembling sometimes improves results and that increasing $k$ for the small SI-Flights dataset harms results.
We provide example outputs of ActiveAED in Appendix~\ref{app:outputs}.

\section{Conclusion}
We have proposed ActiveAED, an AED method that includes human feedback in its prediction loop.
While the proposed approach could be used with every ranking-based AED method, we base ActiveAED on the recently proposed AUM score, which we augment with a novel ensembling scheme based on training dynamics.
We evaluate ActiveAED on eight datasets spanning five different tasks and find that it improves results on seven of them, with gains of up to six pp AP.
In future work, we plan on extending ActiveAED to generative models and structured prediction tasks.
Additionally, we want to use ActiveAED to clean benchmark datasets.
We also plan to investigate the reasons for the observed performance gains of ActiveAED, for instance by exploring the role of model capacity and dataset characteristics~\citep{ethayarajhUnderstandingDatasetDifficulty2022}.
Finally, we would like to study the interplay between ActiveAED and human label variation~\citep{jiangInvestigatingReasonsDisagreement2022,plank-2022-problem}.

\FloatBarrier
\section*{Limitations}
A major limitation of ActiveAED is that it is significantly more compute-intensive than other scoring-based AED methods such as AUM or DM.
This is inherent to the proposed method because the ensemble requires training of multiple models and, after receiving human feedback, the full ensemble has to be re-trained.
Also, the ensembling of ActiveAED requires more training runs than training-dynamics-based AED methods.  
However, most model-based methods require a cross-validation scheme~\citep{klieAnnotationErrorDetection2022}.
The ensembling component of ActiveAED is more data-efficient than these approaches, because it makes use of the training dynamics captured during cross-validation instead of discarding them.
A second limitation of this work is that while we chose baselines that performed strongly in \citet{klieAnnotationErrorDetection2022}, they represent only a fraction of the scoring-based AED methods described in the literature.
Finally, our evaluation is limited to a single language model and it would be interesting to investigate how ActiveAED interacts with larger language models than DistilRoBERTa.

\section*{Ethics Statement}
Datasets with fewer annotation errors can improve model training and evaluation.
While this generally seems desirable, it is subject to the same dual-use concerns as the NLP models that are improved with AED methods.
Additionally, using ActiveAED instead of AUM or DM can make the AED results more accurate, but that comes at the expense of a higher runtime. 
This, in turn, leads to increased energy consumption and, depending on the source of the energy, more $\text{CO}_2$ released~\citep{strubellEnergyPolicyConsiderations2019}, which is highly problematic in the face of the climate crisis.

\section*{Acknowledgements}
We thank the reviewers for their constructive feedback which helped to improve the paper.
Many thanks to the members of MaiNLP and NLPNorth for their comments on the paper.
This research is in parts supported by European Research Council (ERC) grant agreement No.\ 101043235.

\bibliography{acl2023}

\begin{thebibliography}{37}
\expandafter\ifx\csname natexlab\endcsname\relax\def\natexlab#1{#1}\fi

\bibitem[{Alt et~al.(2020)Alt, Gabryszak, and
  Hennig}]{altTACREDRevisitedThorough2020}
Christoph Alt, Aleksandra Gabryszak, and Leonhard Hennig. 2020.
\newblock \href {https://doi.org/10.18653/v1/2020.acl-main.142} {{{TACRED
  Revisited}}: {{A Thorough Evaluation}} of the {{TACRED Relation Extraction
  Task}}}.
\newblock In \emph{Proceedings of the 58th {{Annual Meeting}} of the
  {{Association}} for {{Computational Linguistics}}}, pages 1558--1569,
  {Online}. {Association for Computational Linguistics}.

\bibitem[{Amiri et~al.(2018)Amiri, Miller, and
  Savova}]{amiriSpottingSpuriousData2018}
Hadi Amiri, Timothy Miller, and Guergana Savova. 2018.
\newblock \href {https://doi.org/10.18653/v1/N18-1182} {Spotting {{Spurious
  Data}} with {{Neural Networks}}}.
\newblock In \emph{Proceedings of the 2018 {{Conference}} of the {{North
  American Chapter}} of the {{Association}} for {{Computational Linguistics}}:
  {{Human Language Technologies}}, {{Volume}} 1 ({{Long Papers}})}, pages
  2006--2016, {New Orleans, Louisiana}. {Association for Computational
  Linguistics}.

\bibitem[{Chong et~al.(2022)Chong, Hong, and
  Manning}]{chong-etal-2022-detecting}
Derek Chong, Jenny Hong, and Christopher Manning. 2022.
\newblock \href {https://aclanthology.org/2022.emnlp-main.618} {Detecting label
  errors by using pre-trained language models}.
\newblock In \emph{Proceedings of the 2022 Conference on Empirical Methods in
  Natural Language Processing}, pages 9074--9091, Abu Dhabi, United Arab
  Emirates. Association for Computational Linguistics.

\bibitem[{Dickinson and
  Meurers(2003)}]{dickinsonDetectingErrorsPartofSpeech2003}
Markus Dickinson and W.~Detmar Meurers. 2003.
\newblock Detecting {{Errors}} in {{Part-of-Speech Annotation}}.
\newblock In \emph{10th {{Conference}} of the {{European Chapter}} of the
  {{Association}} for {{Computational Linguistics}}}, {Budapest, Hungary}.
  {Association for Computational Linguistics}.

\bibitem[{Ethayarajh et~al.(2022)Ethayarajh, Choi, and
  Swayamdipta}]{ethayarajhUnderstandingDatasetDifficulty2022}
Kawin Ethayarajh, Yejin Choi, and Swabha Swayamdipta. 2022.
\newblock \href {https://doi.org/10.48550/arXiv.2110.08420} {Understanding
  {{Dataset Difficulty}} with \$\textbackslash
  mathcal\{\vphantom\}{{V}}\vphantom\{\}\$-{{Usable Information}}}.

\bibitem[{Grivas et~al.(2020)Grivas, Alex, Grover, Tobin, and
  Whiteley}]{grivasNotCuteStroke2020}
Andreas Grivas, Beatrice Alex, Claire Grover, Richard Tobin, and William
  Whiteley. 2020.
\newblock \href {https://doi.org/10.18653/v1/2020.louhi-1.4} {Not a cute
  stroke: {{Analysis}} of {{Rule-}} and {{Neural Network-based Information
  Extraction Systems}} for {{Brain Radiology Reports}}}.
\newblock In \emph{Proceedings of the 11th {{International Workshop}} on
  {{Health Text Mining}} and {{Information Analysis}}}, pages 24--37, {Online}.
  {Association for Computational Linguistics}.

\bibitem[{Hemphill et~al.(1990)Hemphill, Godfrey, and
  Doddington}]{hemphillATISSpokenLanguage1990}
Charles~T. Hemphill, John~J. Godfrey, and George~R. Doddington. 1990.
\newblock The {{ATIS Spoken Language Systems Pilot Corpus}}.
\newblock In \emph{Speech and {{Natural Language}}: {{Proceedings}} of a
  {{Workshop Held}} at {{Hidden Valley}}, {{Pennsylvania}}, {{June}}
  24-27,1990}.

\bibitem[{Huang et~al.(2019)Huang, Qu, Jia, and
  Zhao}]{huangO2UNetSimpleNoisy2019}
Jinchi Huang, Lie Qu, Rongfei Jia, and Binqiang Zhao. 2019.
\newblock \href {https://doi.org/10.1109/ICCV.2019.00342} {{{O2U-Net}}: {{A
  Simple Noisy Label Detection Approach}} for {{Deep Neural Networks}}}.
\newblock In \emph{2019 {{IEEE}}/{{CVF International Conference}} on {{Computer
  Vision}} ({{ICCV}})}, pages 3325--3333.

\bibitem[{Jiang and
  de~Marneffe(2022)}]{jiangInvestigatingReasonsDisagreement2022}
Nan-Jiang Jiang and Marie-Catherine de~Marneffe. 2022.
\newblock \href {https://doi.org/10.1162/tacl_a_00523} {Investigating
  {{Reasons}} for {{Disagreement}} in {{Natural Language Inference}}}.
\newblock \emph{Transactions of the Association for Computational Linguistics},
  10:1357--1374.

\bibitem[{Kingma and Ba(2015)}]{kingmaAdamMethodStochastic2014}
Diederik~P. Kingma and Jimmy Ba. 2015.
\newblock Adam: {{A}} method for stochastic optimization.
\newblock In \emph{3rd International Conference on Learning Representations,
  {{ICLR}} 2015, San Diego, {{CA}}, {{USA}}, May 7-9, 2015, Conference Track
  Proceedings}.

\bibitem[{Klie et~al.(2022)Klie, Webber, and
  Gurevych}]{klieAnnotationErrorDetection2022}
Jan-Christoph Klie, Bonnie Webber, and Iryna Gurevych. 2022.
\newblock \href {https://doi.org/10.1162/coli_a_00464} {Annotation {{Error
  Detection}}: {{Analyzing}} the {{Past}} and {{Present}} for a {{More Coherent
  Future}}}.
\newblock \emph{Computational Linguistics}, pages 1--42.

\bibitem[{Kv{\u e}to{\v n} and Oliva(2002)}]{kvetonSemiAutomaticDetection2002}
Pavel Kv{\u e}to{\v n} and Karel Oliva. 2002.
\newblock ({{Semi-}}){{Automatic Detection}} of {{Errors}} in {{PoS-Tagged
  Corpora}}.
\newblock In \emph{{{COLING}} 2002: {{The}} 19th {{International Conference}}
  on {{Computational Linguistics}}}.

\bibitem[{Lacoste et~al.(2019)Lacoste, Luccioni, Schmidt, and
  Dandres}]{lacoste2019quantifying}
Alexandre Lacoste, Alexandra Luccioni, Victor Schmidt, and Thomas Dandres.
  2019.
\newblock \href {http://arxiv.org/abs/1910.09700} {Quantifying the carbon
  emissions of machine learning}.
\newblock \emph{arXiv preprint arXiv:1910.09700}.

\bibitem[{Larson et~al.(2020)Larson, Cheung, Mahendran, Leach, and
  Kummerfeld}]{larsonInconsistenciesCrowdsourcedSlotFilling2020a}
Stefan Larson, Adrian Cheung, Anish Mahendran, Kevin Leach, and Jonathan~K.
  Kummerfeld. 2020.
\newblock \href {https://doi.org/10.18653/v1/2020.coling-main.442}
  {Inconsistencies in {{Crowdsourced Slot-Filling Annotations}}: {{A Typology}}
  and {{Identification Methods}}}.
\newblock In \emph{Proceedings of the 28th {{International Conference}} on
  {{Computational Linguistics}}}, pages 5035--5046, {Barcelona, Spain
  (Online)}. {International Committee on Computational Linguistics}.

\bibitem[{Larson et~al.(2019)Larson, Mahendran, Lee, Kummerfeld, Hill,
  Laurenzano, Hauswald, Tang, and Mars}]{larsonOutlierDetectionImproved2019}
Stefan Larson, Anish Mahendran, Andrew Lee, Jonathan~K. Kummerfeld, Parker
  Hill, Michael~A. Laurenzano, Johann Hauswald, Lingjia Tang, and Jason Mars.
  2019.
\newblock \href {https://doi.org/10.18653/v1/N19-1051} {Outlier {{Detection}}
  for {{Improved Data Quality}} and {{Diversity}} in {{Dialog Systems}}}.
\newblock In \emph{Proceedings of the 2019 {{Conference}} of the {{North
  American Chapter}} of the {{Association}} for {{Computational Linguistics}}:
  {{Human Language Technologies}}, {{Volume}} 1 ({{Long}} and {{Short
  Papers}})}, pages 517--527, {Minneapolis, Minnesota}. {Association for
  Computational Linguistics}.

\bibitem[{Maas et~al.(2011)Maas, Daly, Pham, Huang, Ng, and
  Potts}]{maas-EtAl:2011:ACL-HLT2011}
Andrew~L. Maas, Raymond~E. Daly, Peter~T. Pham, Dan Huang, Andrew~Y. Ng, and
  Christopher Potts. 2011.
\newblock \href {http://www.aclweb.org/anthology/P11-1015} {Learning word
  vectors for sentiment analysis}.
\newblock In \emph{Proceedings of the 49th Annual Meeting of the Association
  for Computational Linguistics: Human Language Technologies}, pages 142--150,
  Portland, Oregon, USA. Association for Computational Linguistics.

\bibitem[{Moro and Lonza(2018)}]{moroElectricityCarbonIntensity2018}
Alberto Moro and Laura Lonza. 2018.
\newblock \href {https://doi.org/10.1016/j.trd.2017.07.012} {Electricity carbon
  intensity in {{European Member States}}: {{Impacts}} on {{GHG}} emissions of
  electric vehicles}.
\newblock \emph{Transportation Research Part D: Transport and Environment},
  64:5--14.

\bibitem[{Northcutt et~al.(2021)Northcutt, Athalye, and
  Mueller}]{northcuttPervasiveLabelErrors2021}
Curtis Northcutt, Anish Athalye, and Jonas Mueller. 2021.
\newblock Pervasive {{Label Errors}} in {{Test Sets Destabilize Machine
  Learning Benchmarks}}.
\newblock \emph{Proceedings of the Neural Information Processing Systems Track
  on Datasets and Benchmarks}, 1.

\bibitem[{Ovadia et~al.(2019)Ovadia, Fertig, Ren, Nado, Sculley, Nowozin,
  Dillon, Lakshminarayanan, and Snoek}]{ovadiaCanYouTrust2019}
Yaniv Ovadia, Emily Fertig, Jie Ren, Zachary Nado, D.~Sculley, Sebastian
  Nowozin, Joshua Dillon, Balaji Lakshminarayanan, and Jasper Snoek. 2019.
\newblock Can you trust your model' s uncertainty? {{Evaluating}} predictive
  uncertainty under dataset shift.
\newblock In \emph{Advances in {{Neural Information Processing Systems}}},
  volume~32. {Curran Associates, Inc.}

\bibitem[{Pedregosa et~al.(2011)Pedregosa, Varoquaux, Gramfort, Michel,
  Thirion, Grisel, Blondel, Prettenhofer, Weiss, Dubourg, Vanderplas, Passos,
  Cournapeau, Brucher, Perrot, and Duchesnay}]{scikit-learn}
F.~Pedregosa, G.~Varoquaux, A.~Gramfort, V.~Michel, B.~Thirion, O.~Grisel,
  M.~Blondel, P.~Prettenhofer, R.~Weiss, V.~Dubourg, J.~Vanderplas, A.~Passos,
  D.~Cournapeau, M.~Brucher, M.~Perrot, and E.~Duchesnay. 2011.
\newblock Scikit-learn: {{Machine}} learning in {{Python}}.
\newblock \emph{Journal of Machine Learning Research}, 12:2825--2830.

\bibitem[{Plank(2022)}]{plank-2022-problem}
Barbara Plank. 2022.
\newblock \href {https://aclanthology.org/2022.emnlp-main.731} {The
  {``}problem{''} of human label variation: On ground truth in data, modeling
  and evaluation}.
\newblock In \emph{Proceedings of the 2022 Conference on Empirical Methods in
  Natural Language Processing}, pages 10671--10682, Abu Dhabi, United Arab
  Emirates. Association for Computational Linguistics.

\bibitem[{Plank et~al.(2014)Plank, Hovy, and
  S{\o}gaard}]{plankLinguisticallyDebatableJust2014}
Barbara Plank, Dirk Hovy, and Anders S{\o}gaard. 2014.
\newblock \href {https://doi.org/10.3115/v1/P14-2083} {Linguistically debatable
  or just plain wrong?}
\newblock In \emph{Proceedings of the 52nd {{Annual Meeting}} of the
  {{Association}} for {{Computational Linguistics}} ({{Volume}} 2: {{Short
  Papers}})}, pages 507--511, {Baltimore, Maryland}. {Association for
  Computational Linguistics}.

\bibitem[{Pleiss et~al.(2020)Pleiss, Zhang, Elenberg, and
  Weinberger}]{pleissIdentifyingMislabeledData2020}
Geoff Pleiss, Tianyi Zhang, Ethan Elenberg, and Kilian~Q Weinberger. 2020.
\newblock Identifying {{Mislabeled Data}} using the {{Area Under}} the {{Margin
  Ranking}}.
\newblock In \emph{Advances in {{Neural Information Processing Systems}}},
  volume~33, pages 17044--17056. {Curran Associates, Inc.}

\bibitem[{Reiss et~al.(2020)Reiss, Xu, Cutler, Muthuraman, and
  Eichenberger}]{reissIdentifyingIncorrectLabels2020}
Frederick Reiss, Hong Xu, Bryan Cutler, Karthik Muthuraman, and Zachary
  Eichenberger. 2020.
\newblock \href {https://doi.org/10.18653/v1/2020.conll-1.16} {Identifying
  {{Incorrect Labels}} in the {{CoNLL-2003 Corpus}}}.
\newblock In \emph{Proceedings of the 24th {{Conference}} on {{Computational
  Natural Language Learning}}}, pages 215--226, {Online}. {Association for
  Computational Linguistics}.

\bibitem[{Sanh et~al.(2020)Sanh, Debut, Chaumond, and
  Wolf}]{sanhDistilBERTDistilledVersion2020}
Victor Sanh, Lysandre Debut, Julien Chaumond, and Thomas Wolf. 2020.
\newblock \href {https://doi.org/10.48550/arXiv.1910.01108} {{{DistilBERT}}, a
  distilled version of {{BERT}}: Smaller, faster, cheaper and lighter}.

\bibitem[{Settles(2012)}]{settlesActiveLearning2012}
Burr Settles. 2012.
\newblock \href {https://doi.org/10.2200/S00429ED1V01Y201207AIM018} {Active
  {{Learning}}}.
\newblock \emph{Synthesis Lectures on Artificial Intelligence and Machine
  Learning}, 6(1):1--114.

\bibitem[{Siddiqui et~al.(2022)Siddiqui, Rajkumar, Maharaj, Krueger, and
  Hooker}]{siddiquiMetadataArchaeologyUnearthing2022}
Shoaib~Ahmed Siddiqui, Nitarshan Rajkumar, Tegan Maharaj, David Krueger, and
  Sara Hooker. 2022.
\newblock \href {https://doi.org/10.48550/arXiv.2209.10015} {Metadata
  {{Archaeology}}: {{Unearthing Data Subsets}} by {{Leveraging Training
  Dynamics}}}.

\bibitem[{Socher et~al.(2013)Socher, Perelygin, Wu, Chuang, Manning, Ng, and
  Potts}]{socher-etal-2013-recursive}
Richard Socher, Alex Perelygin, Jean Wu, Jason Chuang, Christopher~D. Manning,
  Andrew Ng, and Christopher Potts. 2013.
\newblock \href {https://aclanthology.org/D13-1170} {Recursive deep models for
  semantic compositionality over a sentiment treebank}.
\newblock In \emph{Proceedings of the 2013 Conference on Empirical Methods in
  Natural Language Processing}, pages 1631--1642, Seattle, Washington, USA.
  Association for Computational Linguistics.

\bibitem[{Stenetorp et~al.(2012)Stenetorp, Pyysalo, Topi{\'c}, Ohta, Ananiadou,
  and Tsujii}]{stenetorpBRATWebbasedTool2012}
Pontus Stenetorp, Sampo Pyysalo, Goran Topi{\'c}, Tomoko Ohta, Sophia
  Ananiadou, and Jun'ichi Tsujii. 2012.
\newblock {{BRAT}}: A web-based tool for {{NLP-assisted}} text annotation.
\newblock In \emph{Proceedings of the {{Demonstrations}} at the 13th
  {{Conference}} of the {{European Chapter}} of the {{Association}} for
  {{Computational Linguistics}}}, pages 102--107.

\bibitem[{Strubell et~al.(2019)Strubell, Ganesh, and
  McCallum}]{strubellEnergyPolicyConsiderations2019}
Emma Strubell, Ananya Ganesh, and Andrew McCallum. 2019.
\newblock \href {https://doi.org/10.18653/v1/P19-1355} {Energy and {{Policy
  Considerations}} for {{Deep Learning}} in {{NLP}}}.
\newblock In \emph{Proceedings of the 57th {{Annual Meeting}} of the
  {{Association}} for {{Computational Linguistics}}}, pages 3645--3650,
  {Florence, Italy}. {Association for Computational Linguistics}.

\bibitem[{Swayamdipta et~al.(2020)Swayamdipta, Schwartz, Lourie, Wang,
  Hajishirzi, Smith, and Choi}]{swayamdiptaDatasetCartographyMapping2020}
Swabha Swayamdipta, Roy Schwartz, Nicholas Lourie, Yizhong Wang, Hannaneh
  Hajishirzi, Noah~A. Smith, and Yejin Choi. 2020.
\newblock \href {https://doi.org/10.18653/v1/2020.emnlp-main.746} {Dataset
  {{Cartography}}: {{Mapping}} and {{Diagnosing Datasets}} with {{Training
  Dynamics}}}.
\newblock In \emph{Proceedings of the 2020 {{Conference}} on {{Empirical
  Methods}} in {{Natural Language Processing}} ({{EMNLP}})}, pages 9275--9293,
  {Online}. {Association for Computational Linguistics}.

\bibitem[{Tjong Kim~Sang and
  De~Meulder(2003)}]{tjongkimsangIntroductionCoNLL2003Shared2003}
Erik~F. Tjong Kim~Sang and Fien De~Meulder. 2003.
\newblock Introduction to the {{CoNLL-2003 Shared Task}}:
  {{Language-Independent Named Entity Recognition}}.
\newblock In \emph{Proceedings of the {{Seventh Conference}} on {{Natural
  Language Learning}} at {{HLT-NAACL}} 2003}, pages 142--147.

\bibitem[{Varshney et~al.(2022)Varshney, Mishra, and
  Baral}]{varshneyILDAEInstanceLevelDifficulty2022}
Neeraj Varshney, Swaroop Mishra, and Chitta Baral. 2022.
\newblock \href {https://doi.org/10.18653/v1/2022.acl-long.240} {{{ILDAE}}:
  {{Instance-Level Difficulty Analysis}} of {{Evaluation Data}}}.
\newblock In \emph{Proceedings of the 60th {{Annual Meeting}} of the
  {{Association}} for {{Computational Linguistics}} ({{Volume}} 1: {{Long
  Papers}})}, pages 3412--3425, {Dublin, Ireland}. {Association for
  Computational Linguistics}.

\bibitem[{Vlachos(2006)}]{vlachosActiveAnnotation2006}
Andreas Vlachos. 2006.
\newblock Active {{Annotation}}.
\newblock In \emph{Proceedings of the {{Workshop}} on {{Adaptive Text
  Extraction}} and {{Mining}} ({{ATEM}} 2006)}.

\bibitem[{{Yaghoub-Zadeh-Fard} et~al.(2019){Yaghoub-Zadeh-Fard}, Benatallah,
  Chai~Barukh, and Zamanirad}]{yaghoub-zadeh-fardStudyIncorrectParaphrases2019}
Mohammad-Ali {Yaghoub-Zadeh-Fard}, Boualem Benatallah, Moshe Chai~Barukh, and
  Shayan Zamanirad. 2019.
\newblock \href {https://doi.org/10.18653/v1/N19-1026} {A {{Study}} of
  {{Incorrect Paraphrases}} in {{Crowdsourced User Utterances}}}.
\newblock In \emph{Proceedings of the 2019 {{Conference}} of the {{North
  American Chapter}} of the {{Association}} for {{Computational Linguistics}}:
  {{Human Language Technologies}}, {{Volume}} 1 ({{Long}} and {{Short
  Papers}})}, pages 295--306, {Minneapolis, Minnesota}. {Association for
  Computational Linguistics}.

\bibitem[{Zeldes(2017)}]{Zeldes2017}
Amir Zeldes. 2017.
\newblock \href {https://doi.org/http://dx.doi.org/10.1007/s10579-016-9343-x}
  {The {{GUM}} corpus: {{Creating}} multilayer resources in the classroom}.
\newblock \emph{Language Resources and Evaluation}, 51(3):581--612.

\bibitem[{Zhang et~al.(2022)Zhang, Gong, Liu, He, Chen, and
  Zhou}]{zhangALLSHActiveLearning2022}
Shujian Zhang, Chengyue Gong, Xingchao Liu, Pengcheng He, Weizhu Chen, and
  Mingyuan Zhou. 2022.
\newblock \href {https://doi.org/10.48550/arXiv.2205.04980} {{{ALLSH}}:
  {{Active Learning Guided}} by {{Local Sensitivity}} and {{Hardness}}}.

\end{thebibliography}
\bibliographystyle{acl_natbib}

\appendix

\section{Datasets}
\label{app:datasets}
\begin{table*}[h]
    \footnotesize
    \centering
      \begin{tabular}{llllllllll}
        \toprule
        & $|\mathcal{I}|$ & $|\mathcal{I}_\epsilon|$ & $\frac{|\mathcal{I}_\epsilon|}{|\mathcal{I}|}\%$ & $|\mathcal{A}|$ & $|\mathcal{A}_\epsilon|$ & $\frac{|\mathcal{A}_\epsilon|}{|\mathcal{A}|}\%$ & Datasets URL & License \\ 
        \midrule
      ATIS & 4978  & 238   & 4.8   & 4978  & 238   & 4.8   & \href{https://huggingface.co/datasets/mainlp/aed_atis}{\nolinkurl{mainlp/aed_atis}} & LDC\\
      IMDb & 25,000 & 725   & 2.9   & 25000 & 725   & 2.9   & \href{https://huggingface.co/datasets/mainlp/pervasive_imdb}{\nolinkurl{mainlp/pervasive_imdb}} & GPL3 \\
      SST & 8544  & 427   & 5.0   & 8544  & 427   & 5.0   & \href{https://huggingface.co/datasets/mainlp/aed_sst}{\nolinkurl{mainlp/aed_sst}} & unknown\\
      GUM & 1117  & 552   & 49.4  & 13480 & 929   & 6.9   & \href{https://huggingface.co/datasets/mainlp/aed_gum}{\nolinkurl{mainlp/aed_gum}} & \href{https://github.com/amir-zeldes/gum/blob/master/LICENSE.txt}{\nolinkurl{Online}}\\
      CoNLL-2003 & 18,463 & 761   & 4.1   & 13870 & 1133  & 8.2   & \href{https://huggingface.co/datasets/mainlp/aed_conll}{\nolinkurl{mainlp/aed_conll}} & \href{https://huggingface.co/datasets/conll2003#licensing-information}{\nolinkurl{Online}}\\
      SI-Companies & 500   & 454   & 90.8  & 7310  & 1650  & 22.6  & \href{https://huggingface.co/datasets/mainlp/inconsistencies_companies}{\nolinkurl{mainlp/inconsistencies_companies}} & CC-BY4\\
      SI-Flights & 500   & 224   & 44.8  & 2571  & 420   & 16.3  & \href{https://huggingface.co/datasets/mainlp/aed_atis}{\nolinkurl{mainlp/aed_atis}} & CC-BY4\\
      SI-Forex & 520   & 143   & 27.5  & 1632  & 326   & 20.0  & \href{https://huggingface.co/datasets/mainlp/inconsistencies_forex}{\nolinkurl{mainlp/inconsistencies_forex}} & CC-BY4\\
      \bottomrule
      \end{tabular}%
    \caption{Statistics for all datasets that we used in this study. $|\mathcal{I}|$ is the number of total instances, whereas $|\mathcal{I}_\epsilon|$ is the number of instances with at least one wrong annotation. $|\mathcal{A}|$ is the number of total annotations and $|\mathcal{A}_\epsilon|$ the number of erroneous annotations. Datasets URL is the name of our implementation of the dataset in the huggingface datasets hub, which allows to deterministically reproduce all datasets.}
    \label{tab:dataset_statistics}%
  \end{table*}%

Table~\ref{tab:dataset_statistics} lists statistics for all datasets that we used in this work, together with links to the HuggingFace datasets implementations we provide.

\section{Hyperparameters}
\label{app:hyperparameters}
As base model, we choose the 82M parameter model DistilRoBERTa-base\footnote{\url{https://huggingface.co/distilroberta-base}}~\citep{sanhDistilBERTDistilledVersion2020}, which is licensed under apache-2.0.
In all experiments, we perform 10-fold cross validation.
We manually optimize the hyperparameters of ActiveAED on ATIS and SI Flights, resulting in a learning rate of 5e-5 and a batch size of 64.
We adapt the number of epochs to the size of the dataset: for the SI datasets, we set it to 40, for ATIS to 20, for GUM, CoNLL and SST to 10, and for IMDb to 5 and use Adam~\citep{kingmaAdamMethodStochastic2014}.
We set the number of instances that the annotator corrects in a single pass $k$ to 50 for all datasets because this is small enough so that an annotator can handle it in a single annotation session but large enough that gains could be observed after a single iteration on SI Flights.

\section{Further Ablation Studies}
\label{app:ablation}
\begin{table}[htbp]
    \centering
      \begin{tabular}{lll}
    \toprule
    &   ATIS & SI-Flights \\
      \midrule
      ActiveAED & 98.6±0.1 & 86.6±0.5 \\
      \midrule
      w/o active & 98.7±0.1 & \textbf{80.3±0.6} \\
      w/o test ens. & \textbf{98.3±0.1} & \textbf{84.3±0.5} \\
      w/o train ens. & \textbf{97.4±0.3} & 89.2±1.2 \\
      k = 100 & \textbf{98.5±0.1} & \textbf{86.4±0.5} \\
      k = 200 & 98.7±0.0 & \textbf{82.7±0.7} \\
      \bottomrule
      \end{tabular}%
    \caption{Results of the ablation study. All modifications denote independent changes to ActiveAED. I.e. 'w/o test ens.' is ActiveAED without the test ensembling but with train ensembling and the human-in-the-loop component. Scores with a lower average than that of ActiveAED are in bold.}
    \label{tab:ablation}%
  \end{table}%

Table~\ref{tab:ablation} gives results for our full ablation study.
We find that for ATIS, where ensembling was the main driver of improved results, ablating both train and test ensembling leads to worse results.
For SI-Flights, the variant without test ensembling leads to worse results, whereas omitting train ensembling improves results.
We hypothesized that, for small datasets, increasing $k$ would lead to worse results.
Our results confirm this.
Setting $k = 100$ leaves results almost unchanged, whereas $k = 200$ leads to a dramatic drop of $3.9$ pp AP on SI-Flights, without affecting performance on the much larger ATIS.

\section{Compute Resources for Experiments}
We estimate the total computational cost of our experiments including development of the method to be around 1000 GPU hours on an 80GB A100.
As per the ML $\text{CO}_2$ Impact tool~\citep{lacoste2019quantifying}\footnote{\url{https://mlco2.github.io/impact/}} and an average carbon intensity of electricity for Germany of $0.485 \frac{\text{kg}}{\text{kWh}} \text{CO}_2$~\citep{moroElectricityCarbonIntensity2018}  this amounts to roughly 121 kg $\text{CO}_2$ emitted.

\section{Example Outputs}
\label{app:outputs}
\begin{figure*}
  \centering
  \includegraphics[width=0.8\linewidth]{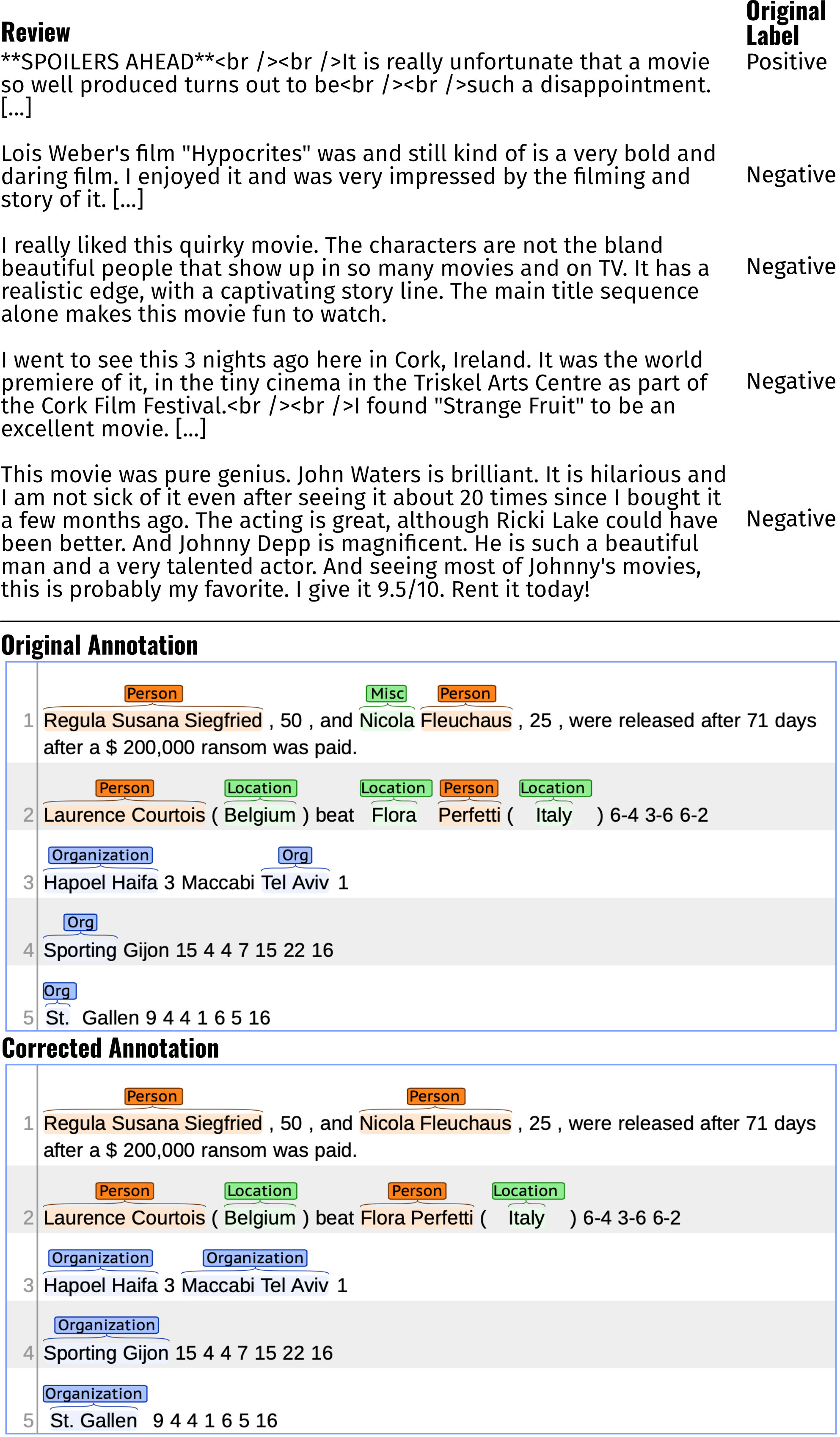}
  \caption{Five instances with highest error scores assigned by ActiveAED for IMDb (top) and CONLL-2003 (bottom; visualized with brat~\citep{stenetorpBRATWebbasedTool2012}). All original annotations are erroneous.}
  \label{fig:example_errors}
\end{figure*}

Example outputs for IMDb and CONLL-2003 can be found in Figure Appendix~\ref{fig:example_errors}. 
We show the five instances with the highest error scores assigned by ActiveAED.
All instances contain an annotation error.

\end{document}